\documentclass[conference]{IEEEtran}
\IEEEoverridecommandlockouts
\usepackage{cite}
\usepackage{amsmath,amssymb,amsfonts}
\usepackage{algorithmic}
\usepackage{graphicx}
\usepackage{textcomp}
\usepackage{xcolor}
\usepackage{comment}
\usepackage{url}
\usepackage{booktabs}
\usepackage[hidelinks,breaklinks]{hyperref}
\usepackage{orcidlink}
\hypersetup{
  hidelinks,
  pdftitle={Temporal Conditioning for Human Behavioral Cloning in Chess},
  pdfauthor={Victor Dziwniel and Tianxiang Lu}
}
\usepackage{eso-pic}
\AddToShipoutPictureBG*{%
  \AtPageUpperLeft{%
    \put(\LenToUnit{0.5\paperwidth},\LenToUnit{-0.18in}){%
      \makebox[0pt][c]{\parbox[t]{0.86\paperwidth}{\centering\scriptsize
        \copyright~2026 IEEE. Personal use of this material is permitted. Permission from IEEE must be obtained for all other uses, in any current or future media, including reprinting/republishing this material for advertising or promotional purposes, creating new collective works, for resale or redistribution to servers or lists, or reuse of any copyrighted component of this work in other works.\\
        Accepted version of: V.~Dziwniel and T.~Lu, ``Temporal Conditioning for Human Behavioral Cloning in Chess,'' 2026 IEEE Gaming, Entertainment, and Media Conference (GEM), Berlin, Germany, 2026, pp.~1--6. \url{https://doi.org/10.1109/GEM72276.2026.11747060}}}}}}
\def\BibTeX{{\rm B\kern-.05em{\sc i\kern-.025em b}\kern-.08em
    T\kern-.1667em\lower.7ex\hbox{E}\kern-.125emX}}
\begin{document}

\title{Temporal Conditioning for Human Behavioral Cloning in Chess}

\author{
\IEEEauthorblockN{Victor Dziwniel and 
Tianxiang Lu\,\orcidlink{0009-0006-4462-1636}}
\IEEEauthorblockA{
IU International University of Applied Sciences, Erfurt, Germany\\
Email: victor.dziwniel@iu-study.org, tianxiang.lu@iu.org
}
}

\maketitle

\begin{abstract}
Behavioral cloning models in chess typically predict human moves from board states while ignoring temporal inputs such as the remaining clock. However, time pressure is a major determinant of human decision-making quality, and temporal state is a missing dimension of human behavioral modeling.

This paper introduces a time-aware transformer that extends a Chessformer-style architecture with learned temporal conditioning tokens encoding player clock state, time control, and recent tempo, together with behavioral heads for thinking time and resignation. The model is trained on large-scale online chess data and evaluated against time-agnostic baselines and the state of the art.

Results show that temporal conditioning yields modest overall accuracy improvements but substantially stronger gains under time pressure. Beyond move choice, it also produces more accurate thinking-time and resignation predictions. More importantly, the model better matches human move-quality degradation in critical time regimes, reducing the systematic over-strength observed in time-agnostic behavioral cloning models.

These results suggest that temporal state is a previously underutilized but important signal for human behavioral cloning in chess.
\end{abstract}


\section{Introduction}
Most chess engines are built to play as strongly as possible. A different goal is to play like a human: reproducing the distribution of moves a player of a given strength would choose, including their mistakes. Such models power realistic sparring partners and bots. They also drive coaching tools that show a learner the kind of move a player at their level would make, and cheat-detection systems that flag games too strong to be human. From a player's perspective, believability matters as much as raw accuracy, an opponent that feels human keeps players engaged, while one that plays robotically breaks the experience.

The behavioral cloning paradigm in chess was first established by Maia \cite{maia} with per-rating convolutional networks, which were later unified into a single skill-conditioned model \cite{maia2}; Allie \cite{allie} moved to a text-based transformer over the move transcript, with search; and Maia-3 \cite{maia3}, built on the Chessformer architecture, returned to a spatial representation, treating the 64 squares as tokens in an encoder-only transformer. However, each of them discards the clock. Remaining time does not condition the policy, so none of these models can learn how a human's move choices shift as time runs out.

Research on the speed--accuracy trade-off in chess shows that time restriction shifts the balance between recognition-based and calculation-based play, changing which moves a human is likely to choose \cite{burns}. Human decision quality falls sharply as the clock runs down, which we quantify in Section~\ref{sec:counterfactual}. A model blind to the clock is forced to learn a single averaged policy that conflates comfortable and time-pressured play. This averaging is most damaging in critically time-pressured play, where human behavior diverges most.

This paper addresses this gap with one claim: conditioning a strong chess policy on temporal state makes its behavior more human-like, with the effect concentrated under time pressure, where a time-blind model keeps playing too strongly precisely when humans play worst. Our contributions are:
\begin{itemize}
\item input-level temporal conditioning for a chess transformer: clock state, time control, and recent tempo supplied as conditioning tokens the policy attends to;
\item a controlled, step-matched ablation at the 23M scale showing that the accuracy gain concentrates monotonically under time pressure and with player skill, and that the sampled policy reproduces the human move-quality degradation that a time-blind twin cannot express;
\item two intervention probes (counterfactual injection and permutation of the dynamic temporal inputs) that isolate the trained model's dependence on the clock from its dependence on correlated position features; and
\item a graded comparison against the released Maia-3 model, from its own filtered evaluation regime to the full spectrum of human play.
\end{itemize}

\section{Related Work}
\textbf{Human behavioral cloning in chess.} Maia \cite{maia} trained one convolutional network per rating band. Maia-2 \cite{maia2} replaced these with a single architecture conditioned on a skill embedding, enabling smooth interpolation across ratings. Allie \cite{allie} reframed the problem as next-token prediction over game transcripts, gaining access to full history and predicting thinking time and resignation in addition to the move, with KL-regularized search at inference \cite{klsearch}. Maia-3 \cite{maia3} represents the board directly, one token per square, with a geometric attention bias, achieving the strongest reported move-matching accuracy. Related work has also demonstrated grandmaster-level play from transformers without search \cite{ruoss}, a different objective from human alignment, and recent work extends Maia-3 toward individual style with rating-conditioned residual player embeddings \cite{carlson}.

\textbf{Time as a signal.} Large-scale analyses of online play have related human error to player skill, position difficulty, and the time available for a move \cite{anderson}, yet existing models have handled the clock in one of three ways. The Maia line filters time pressure out: training data excludes all positions after the first time a player drops below 30 seconds remaining \cite{maia,maia3}, so the models never see the regime where the clock dominates behavior. Allie retains these positions and includes the static time control in its input, but predicts thinking time only as an output; its policy never observes the dynamic clock state, i.e., remaining time or recent tempo \cite{allie}. In recent work, UniMaia-Aux \cite{unimaia} modulates a frozen chess policy through natural-language prompts whose metadata can include remaining time and last-move delay, with auxiliary delay and resignation objectives. Overall, existing approaches either filter time pressure out or use clock information only indirectly, through auxiliary objectives or prompting rather than integrating dynamic temporal state directly into policy generation.

\section{Method}
\subsection{Backbone}
We build on the Chessformer architecture \cite{maia3}, the encoder-only backbone of the Maia-3 models, and leave it unchanged. Each of the 64 squares is a token whose features are dominated by the board itself: one-hot piece occupancy over the current and seven previous positions, to which a few auxiliary planes (en passant, castling rights, and repetition flags) and the skill embedding are appended as additional channels. Player and opponent skill are encoded by interpolating, independently per side, between two learned anchor embeddings at Elo 0 and 5000; the resulting vector is the same for every square and is concatenated into the per-square channels before a single input projection forms the tokens. Each encoder block applies a geometric attention bias that injects board geometry into attention, together with RMSNorm and a GELU feed-forward sub-layer. A bilinear from/to factorization over the 64 squares produces the policy, and a mean-pooled head predicts the win/draw/loss outcome. Our 23M configuration uses an embedding width of 512, 16 attention heads, and 8 blocks; the 79M configuration doubles the width to 1024 with 32 heads. Everything above is the Chessformer design unchanged; our additions, temporal conditioning and the behavioral heads, follow.

\subsection{Temporal conditioning}
\label{sec:conditioning}
We derive five signals: the base time and increment of the \emph{time control}, the two players' remaining clocks (\emph{clock state}), and a vector summarizing the move (ponder) times of the last eight plies (\emph{recent tempo}). We refer to the three collectively as the model's \emph{temporal state}, and to their encoded form as the model's \emph{temporal inputs}. The clock state and recent tempo vary within a game and form the \emph{dynamic} part of the temporal state, as opposed to the static time control. They are encoded into four conditioning tokens, so that the time control, each player's clock, and the recent tempo remain separable to attention. The four tokens are prepended to the 64 board tokens, giving a sequence of length 68. The two clock tokens interpolate between two learned anchor embeddings on a log-time scale so nearby clock states receive nearby embeddings without quantization. Because the transformer attends to these tokens like any other, the predicted move distribution at each square can depend directly on the clock. A channel-concatenated variant that folds the same five signals into the per-square input projection (injecting time exactly as the backbone encodes player skill, with no change to sequence length) trails the prepended-token form by roughly 0.15 percentage points of validation move-matching accuracy, consistently across training; we therefore report the token form.

\subsection{Behavioral heads}
Beyond the policy and value heads, and following Allie \cite{allie}, which predicts thinking time and resignation as well as moves, the model carries behavioral heads that share the trunk and are trained jointly: a move-conditional thinking-time head that predicts how long a human would deliberate before each candidate move, and a resignation head that predicts whether a human will resign. These let the model express the pacing and concession behavior of a human player.

The thinking-time head predicts a full distribution over discrete one-second bins and conditions it on the move. The head emits a position-level baseline distribution from an attention pool over the 64 square tokens, plus per-square \emph{from} and \emph{to} logits; the distribution for a candidate move is the baseline plus that move's source- and destination-square terms. Every legal move thus gets its own predicted distribution from three small projections, which keeps the head cheap while leaving it move-conditional: the same position yields a fast distribution for a forced recapture and a slow one for a dangerous sacrifice. The bins are one second wide (121 in total, covering zero to 120 seconds, with the top bin absorbing any longer think), which matches the one-second resolution at which the Lichess database exports clocks; treating time as classification lets the head capture the skewed, multi-modal shape of human think times and avoids a regression to the mean.

\section{Experimental Setup}
\label{sec:setup}
\subsection{Data} 
We train on rated Lichess games with clock annotations from 2023 through December 2025 \cite{lichess}. Positions are balanced across player skill (Elo 1200--2500, in 100-point bins) and, within each skill bin, evenly across three time controls: bullet, blitz, and rapid. Including bullet broadens the range of time pressure the model observes and covers the regime where the clock signal matters most. The two extreme time controls, ultra-bullet and classical, are retained but not balanced, capped at roughly one and five percent of the balanced set respectively; positions outside the 1200--2500 band are likewise kept under a cap. The balanced training set contains 1.8B positions. All evaluations use a held-out test set of 1{,}500{,}029 positions from May 2026 games, which postdate every training window in this paper (including Maia-3's). Of these, 1{,}485{,}359 are move positions and 14{,}670 are terminal positions; move-based metrics use only the former. The test set covers the same 1200--2500 rating range as training but keeps the natural time-control mix (49\% blitz, 38\% bullet, 13\% rapid).

\subsection{Models and protocol} 
The core ablation compares two 23M models trained for one million steps each under one protocol: a \emph{time-conditioned} model and a \emph{time-blind} model that receives constant (zeroed) temporal inputs. The models share the same backbone, data, optimization schedule, and evaluated outcome, thinking-time, and resignation heads. We follow Maia-3's size naming; our conditioning tokens and heads add about 0.4M parameters at the small size and 0.7M at the large size (23.3M vs. 22.9M and 79.6M vs. 78.9M, respectively). The conditioned model also carries a small auxiliary time-forfeit head that is degenerate without clock access (66{,}817 parameters, 0.29\%) and is not evaluated in this paper.

Training uses AdamW at an effective batch size of 1024 and peak learning rates of $1.25\times10^{-4}$ and $1\times10^{-4}$ for the 23M and 79M configurations respectively. We additionally run both arms of the ablation with three seeds each for 250{,}000 steps to measure training variance. This sweep is reported in Section~\ref{sec:accuracy}; the headline numbers throughout remain those of the single full-budget run per arm.

For comparison to the state of the art, we additionally train a 79M time-conditioned model and evaluate the official released Maia-3 79M checkpoint \cite{maia3} on the same positions with their own inference code; our harness reproduces Maia-3's published 57.1\% move-matching accuracy on their test data under their evaluation protocol before evaluating the checkpoint on our test set.

\subsection{Metrics} 
\label{sec:metrics}
The primary metric is top-1 move-matching accuracy: the fraction of positions where the model's most likely legal move equals the human move. To assess move \emph{quality}, we compute centipawn loss (CPL) against Stockfish (depth 12) for moves sampled from the policy at temperature 1.0; CPL measures how far a move falls below the engine's best move and lets us compare the model's error profile to the human one. The behavioral heads are evaluated with win/draw/loss accuracy, thinking-time error in seconds, and resignation precision, recall, and F1. Move metrics are reported as a function of time pressure, operationalized as the remaining clock as a fraction of base time and grouped into four buckets: comfortable (above 50\%), moderate (20--50\%), pressured (5--20\%) and critical (below 5\%); the test set contains 47{,}562 critical positions. Ablation deltas are paired per position with bootstrap confidence intervals; system-comparison intervals are additionally clustered by game since positions within a game are correlated. All intervals are over positions.

\subsection{Intervention probes}
\label{sec:probes}
Bucket-level differences alone cannot separate what a model reads from the clock from what it reads off the board, because time-pressure buckets correlate with position type. We therefore use two probes that hold positions fixed and manipulate only the temporal inputs. \emph{Counterfactual injection} re-runs the model on the same positions, overriding the clock state and recent tempo, and scores the sampled moves with the engine. \emph{Permutation} replaces a position's dynamic temporal inputs (clock state and recent tempo) with those of a random position from the same time control, which preserves their distribution but removes their link to the board.

\subsection{Reproducibility}
\label{sec:reproducibility}
Each model is trained on a single NVIDIA RTX 5080 (16\,GB) in PyTorch with fp16 mixed precision, gradient clipping at max norm 1.0, and dropout 0.1, using AdamW (weight decay 0.01) under cosine decay after 2{,}000 linear warmup steps. The joint objective weights the policy, WDL, thinking-time, promotion, resignation, and time-forfeit losses by 1.0, 0.5, 0.5, 0.1, 0.1, and 0.1, respectively; the binary resignation and time-forfeit losses use positive-class weights of 20 and 50. Move quality is scored with Stockfish 16 at depth 12.
The implementation, model configurations, seeds, trained weights, and detailed experimental results are publicly available on GitHub\footnote{\url{https://github.com/Prof-it/temporal-chess-model}}.

\section{Results}
\label{sec:results}
\subsection{Move-matching accuracy concentrates under pressure}
\label{sec:accuracy}
Table~\ref{tab:accuracy} reports top-1 accuracy by time-pressure bucket for the step-matched 23M pair. In aggregate, the time-conditioned model is only $+0.73$~pp ahead, but the paired delta widens monotonically with time pressure, reaching $+2.45$~pp in the critical bucket. With ample time the two models agree on the large majority of positions; the clock signal contributes most when little time remains, which is consistent with the speed--accuracy trade-off \cite{burns}.

A reduced-budget seed sweep shows that the move-matching effect and its pressure-dependent shape are reproducible. Across three matched seed pairs trained for 250{,}000 steps, the mean delta is $+0.61$~pp overall and $+2.36$~pp in the critical bucket. Every time-conditioned run outperforms every time-blind run in both cases, and the monotonic widening from comfortable to critical positions holds in all three matched pairs.

The lower block of Table~\ref{tab:accuracy} splits the delta by player rating. The overall gain also grows monotonically with player skill, with disjoint CIs between the lowest and highest bands, and generally tends to increase with skill within each pressure bucket. The largest gain occurs for the strongest players in critical time trouble. Stronger players plausibly respond to the clock more systematically, with deliberate time management and practiced scramble play, so their time-dependent behavior is more learnable.

\begin{table}[htbp]
\caption{Top-1 move-matching accuracy (\%) by time-pressure bucket,
step-matched 23M models, 1{,}485{,}359 positions. Brackets: paired bootstrap 95\% CIs
over positions. Parentheses: max$-$min spread over the three matched seed pairs trained for 250{,}000 steps.}
\label{tab:accuracy}
\centering
\renewcommand{\arraystretch}{1.2}
\setlength{\tabcolsep}{2pt}

\resizebox{\columnwidth}{!}{%
\begin{tabular}{lccccc}
\hline
\textbf{Model} & \textbf{Overall} & \textbf{Comf.} & \textbf{Mod.}
& \textbf{Press.} & \textbf{Crit.} \\
& & ($>$50\%) & (20--50\%) & (5--20\%) & ($<$5\%) \\
\hline

Time-conditioned & 56.19 & 54.76 & 59.41 & 59.60 & 59.05 \\
Time-blind       & 55.46 & 54.18 & 58.60 & 58.45 & 56.61 \\

\hline
$\Delta$ (pp)
& +0.73
& +0.58
& +0.81
& +1.15
& \textbf{+2.45} \\

95\% CI
& {\scriptsize[.69,.77]}
& {\scriptsize[.52,.63]}
& {\scriptsize[.69,.92]}
& {\scriptsize[1.00,1.31]}
& {\scriptsize[2.14,2.72]} \\

\hline
\multicolumn{6}{l}{\textit{Paired $\Delta$ (pp) by rating band}} \\
\hline

Elo 1200--1500
& +0.61 {\scriptsize[.53,.69]}
& +0.50
& +0.70
& +1.05
& +1.93 \\

Elo 1500--1800
& +0.69 {\scriptsize[.62,.76]}
& +0.55
& +0.76
& +1.07
& +2.54 \\

Elo 1800--2100
& +0.78 {\scriptsize[.70,.87]}
& +0.61
& +0.89
& +1.16
& +2.49 \\

Elo 2100--2500
& +0.91 {\scriptsize[.78,1.03]}
& +0.71
& +0.86
& +1.37
& \textbf{+2.66} \\

\hline
\multicolumn{6}{l}{\textit{Mean across three matched seed pairs}} \\
\hline

Seed sweep
& +0.61 {\scriptsize(.07)}
& +0.46 {\scriptsize(.02)}
& +0.59 {\scriptsize(.05)}
& +1.12 {\scriptsize(.41)}
& \textbf{+2.36} {\scriptsize(.50)} \\

\hline
\end{tabular}%
}
\end{table}

\subsection{Move quality under the clock: alignment and attribution}
\label{sec:counterfactual}
\begin{figure}[htbp]
\centering
\includegraphics[width=\columnwidth]{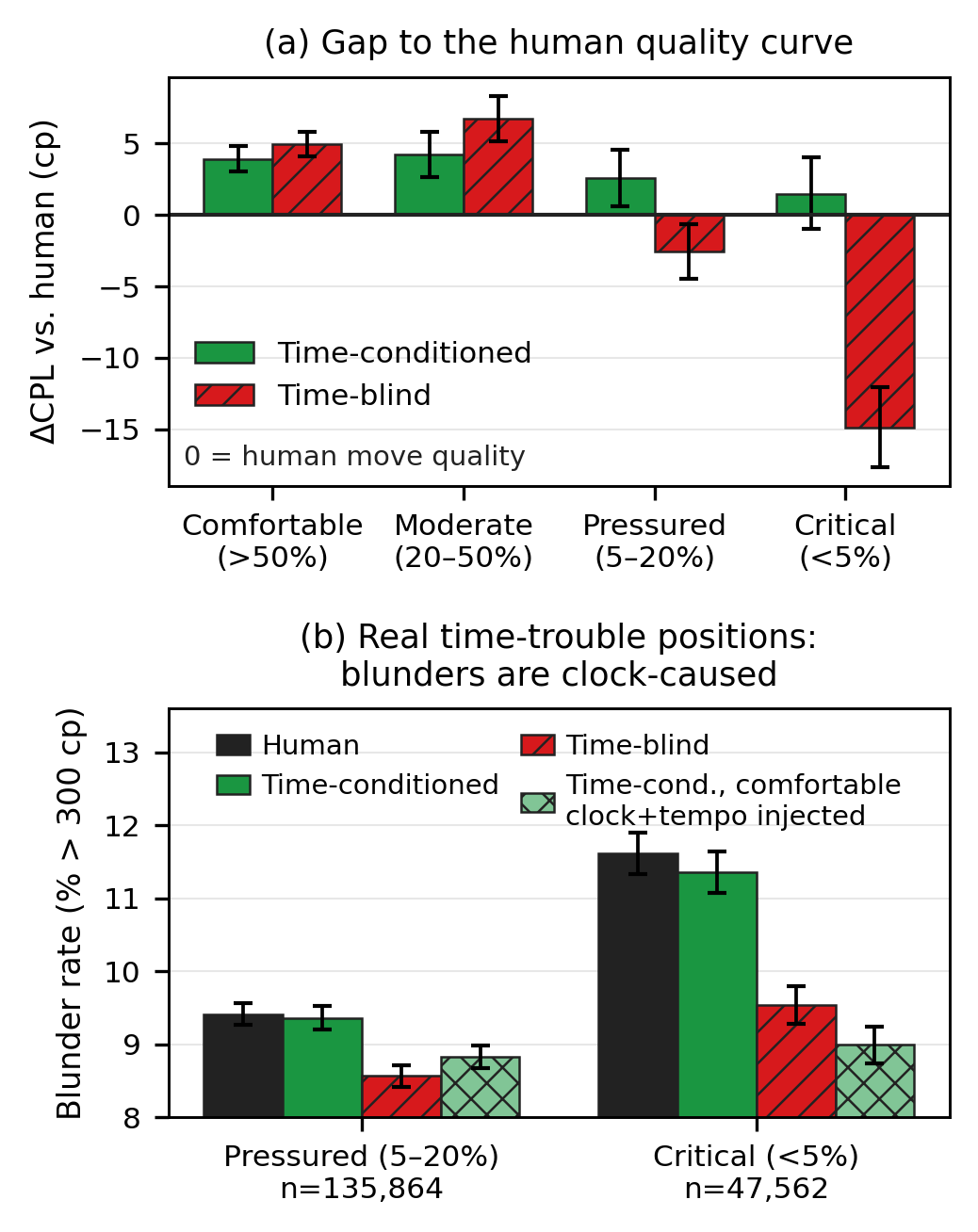}
\caption{Sampled-move quality (temperature 1.0, depth 12 Stockfish, 95\% bootstrap CIs). (a) Deviation from human CPL by time-pressure bucket, paired per position over 329k positions: in critically time-pressured positions, the time-blind model is 14.9 CPL too strong while the time-conditioned model deviates by just +1.4 CPL from the human baseline. (b) Counterfactual on 183{,}426 real time-trouble positions: the time-conditioned model matches the human blunder rate at the real clock; injecting comfortable dynamic temporal inputs (clock state and recent tempo) on the same boards reduces its blunder rate to 9.0\%.}
\label{fig:quality}
\end{figure}

Accuracy alone does not reveal whether a model's mistakes are human-like. Human move quality degrades as the clock runs down: from 60 CPL of sampled-move loss in comfortable positions to 136 CPL in critical ones on our 329k-position quality evaluation, drawn from bullet, blitz, and rapid games and balanced over skill and time pressure. Fig.~\ref{fig:quality}(a) plots each model's deviation from this human curve, paired per position. The time-conditioned policy stays within a few CPL of human quality in every bucket; in the critical bucket its sampled moves land 1.4 CPL above the human mean (CI $[-1.1, +4.3]$). The time-blind policy instead \emph{crosses} the human curve: too weak when time is ample, then too strong once pressure sets in, ending $14.9$ CPL too strong in the critical bucket. This pattern is consistent with a single policy averaged over clock states it cannot distinguish.

The counterfactual probe in Fig.~\ref{fig:quality}(b) holds the board fixed and varies only the temporal inputs (Section~\ref{sec:probes}), on 183{,}426 positions where the human really was in time trouble, so the human blunder rate on the same boards is a true baseline. Humans blunder on 11.6\% of the critical positions, and the time-conditioned model at the real clock closely matches them (11.4\%); the time-blind model, seeing identical boards but no clock, under-blunders at 9.5\%; and injecting comfortable dynamic temporal inputs drops the conditioned model to 9.0\%. The pressured bucket shows the same pattern. On these boards, the temporal inputs alone account for the difference between matching and missing the human blunder rate. Sweeping the injected clock on a separate bucket-representative sample shows a graded increase in blunder rate, from 5.2\% at a full clock to 8.8\% at 2\% of base time.

\subsection{Behavioral human-likeness beyond move choice}
\label{sec:behavioral}
Human-likeness extends beyond the choice of move. Table~\ref{tab:aux} reports the behavioral heads, which the time-blind baseline also carries, trained on the same data but without clock access. The time-conditioned model predicts game outcomes more accurately, with much of the gap coming from time-trouble games whose outcome hinges on the clock race, and predicts resignation with higher precision and F1 (7{,}476 positives, a 0.5\% base rate).

The thinking-time head shows the largest effect of clock access, and the lower block of Table~\ref{tab:aux} shows where it comes from. In critical positions, humans move in 1.2~s on average; the conditioned head predicts 1.3~s, while the time-blind head predicts 3.1~s. The latter cannot know that the player has no time to think, or which time control the game uses, so it predicts the deliberation of an average position. A bot pacing its play with the time-blind head would think too long during time scrambles, a conspicuous tell of non-human play.

We also verify that the head's move-conditionality provides additional predictive signal. Three predictors of human think time are compared on the full test set. Conditioning on the actually played move yields 2.33~s MAE. Replacing this with a policy-weighted average over legal moves increases error to 2.43~s, and ablating move terms entirely further degrades performance to 3.75~s. Log-likelihood ranks the methods identically. Thus, the identity of the played move provides additional predictive information about deliberation time beyond what is available from the position and aggregated move statistics. Players in Allie's Lichess deployment reported that the bot deliberated too long in positions with only one reasonable move \cite{allie}, the case a position-level head must average over.

\begin{table}[htbp]
\caption{Behavioral heads, step-matched 23M models. Lower block: thinking-time prediction by bucket (human mean, mean prediction for each head, MAE; seconds).}
\label{tab:aux}
\centering
\renewcommand{\arraystretch}{1.2}
\setlength{\tabcolsep}{2.5pt}
\begin{tabular}{lccc}
\hline
\textbf{Head (metric)} & & \textbf{Time-cond.} & \textbf{Time-blind} \\
\hline
Outcome, WDL accuracy (\%)  & & \textbf{64.9} & 60.8 \\
Thinking time, MAE (s)      & & \textbf{2.33} & 3.45 \\
Thinking time, median AE (s)& & \textbf{0.94} & 1.91 \\
Resignation, P/R/F1 & & \textbf{.292}/.449/\textbf{.354} & .195/.497/.280 \\
\hline
\textbf{Think time by bucket} & \textbf{human} & pred (MAE) & pred (MAE) \\
\hline
Comfortable ($>$50\%) & 4.5~s & 4.3 (2.55) & 5.2 (3.59) \\
Moderate (20--50\%)   & 4.5~s & 4.3 (2.33) & 5.5 (3.55) \\
Pressured (5--20\%)   & 2.4~s & 2.5 (1.23) & 4.2 (2.63) \\
Critical ($<$5\%)     & 1.2~s & \textbf{1.3 (0.63)} & 3.1 (2.14) \\
\hline
\end{tabular}
\end{table}

\subsection{Comparison with the released state of the art}
\label{sec:sota}
Finally, we compare against the released Maia-3 checkpoint \cite{maia3}, the strongest published human move predictor, on the same 1.49M-position test set, which postdates both models' training data. Maia-3 differs from our 79M model in data, recipe, and training budget as well as temporal conditioning, so this is a comparison of systems and not a second ablation; the permutation probe below does the attribution.

\begin{figure}[htbp]
\centering
\includegraphics[width=\columnwidth]{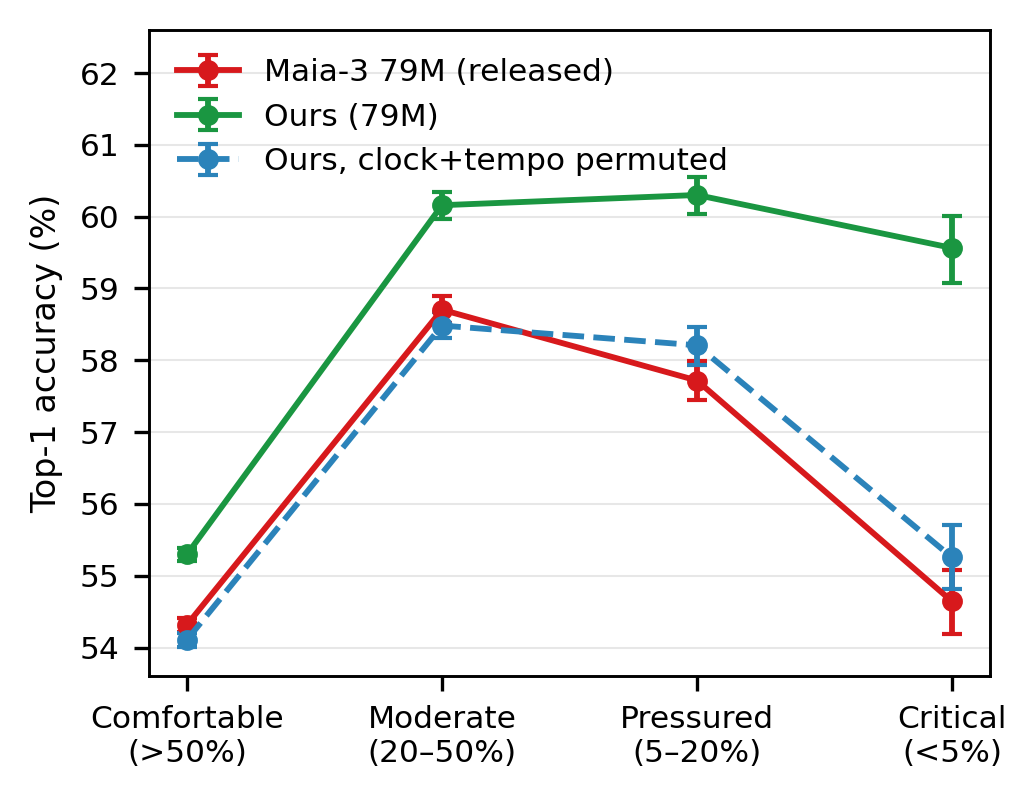}
\caption{Top-1 accuracy by time-pressure bucket: released Maia-3 79M, our time-conditioned 79M, and the same model with its dynamic temporal inputs (clock state and recent tempo) permuted (game-clustered 95\% CIs). Permuting them brings our model close to Maia-3's curve in every bucket.}
\label{fig:maia3}
\end{figure}

Fig.~\ref{fig:maia3} shows the 79M comparison. Our time-conditioned 79M model is $+1.34$~pp ahead overall, and the gap widens with pressure to $+4.92$~pp in critical time trouble. Permuting our model's dynamic temporal inputs erases 4.3 of the 4.9 critical-bucket points.

Because Maia-3 trains only on blitz and discards positions after the first sub-30-second clock \cite{maia3}, evaluation on bullet games and time scrambles extends beyond its training distribution. Table~\ref{tab:ladder} therefore presents a graded comparison, beginning with Maia-3's own evaluation protocol and progressively adding the regimes excluded by its filtering procedure. Under Maia-3's evaluation regime our 79M model is already ahead; the gap grows when blitz time trouble is added back, and more than doubles again on the full set. The per-bucket breakdown in Fig.~\ref{fig:maia3} shows the difference is largest in critically time-pressured positions.

\begin{table}[htbp]
\caption{Graded-fairness comparison with released Maia-3 79M on May-2026 positions, game-clustered 95\% CIs. Each row relaxes one of the Maia-3 evaluation filters.}
\label{tab:ladder}
\centering
\renewcommand{\arraystretch}{1.2}
\setlength{\tabcolsep}{4pt}
\begin{tabular}{lrccc}
\hline
\textbf{Regime} & \textbf{n} & \textbf{Maia-3} & \textbf{Ours} & $\Delta$ (pp) \\
\hline
Maia-3 protocol$^{\mathrm{a}}$ & 515{,}927 & 57.46 & 57.88 & +0.42 {\scriptsize[.35,.50]} \\
Blitz, unfiltered & 721{,}907 & 57.04 & 57.60 & +0.56 {\scriptsize[.50,.62]} \\
Full set & 1{,}485{,}359 & 55.45 & 56.79 & +1.34 {\scriptsize[1.29,1.39]} \\
\hline
\multicolumn{5}{l}{{$^{\mathrm{a}}$blitz only, plies $\geq$ 10, positions before either clock is $<$30~s.}}
\end{tabular}
\end{table}

\section{Limitations}
We evaluate human-likeness offline, through behavioral alignment metrics; we have not tested it in human--bot interaction. Improved alignment in move and timing distributions may not translate directly into perceived realism, which can depend on salient interaction-level cues. Testing this in a blinded user study remains future work.

The controlled ablation covers one scale (23M), with three seeds per arm only at the reduced budget; the seed sweep therefore bounds run-to-run variation on the move-matching delta, but the quality, counterfactual, and behavioral-head results rest on one run per arm. The 79M model has no time-blind twin, so its ablation evidence is the permutation probe. The Lichess database exports clocks at one-second granularity, so think times below one second are quantized away; this makes the think-time targets noisiest where moves are fastest (the critical bucket, mean 1.2~s). The time-blind model omits the time-forfeit head (clock-degenerate), a 0.29\% parameter difference. All data are from Lichess, and our results describe models fitted to that data rather than human behavior itself; behavior on other platforms or over-the-board play may differ.

\section{Conclusion}
Human play in chess is shaped by the clock, but leading behavioral cloning models have not conditioned their policies directly on dynamic temporal state: prior models either filter time pressure out of training or treat move timing as a prediction target. Our model conditions the policy of a Chessformer backbone on temporal state through four learned tokens and adds move-conditional thinking-time and resignation heads. 

In a step-matched ablation differing in temporal conditioning, with the move-matching pattern replicated across three reduced-budget seeds per arm, the clock signal yields a modest overall accuracy gain that more than triples in critical time trouble; it lets the sampled policy reproduce the human move-quality degradation that a time-blind model systematically underestimates by 15 centipawns; and interventions on the temporal inputs alone move the model between matching and missing the human blunder rate on fixed boards. Separately, a 79M time-conditioned model exceeds the released state of the art by almost five accuracy points in critical time trouble; data and recipe differ there too, but permuting the dynamic temporal inputs erases nearly all of that margin.

For the applications that motivate human modeling (believable opponents, coaching tools, cheat detection), the time-trouble regime is a routine feature of online play, and a model that cannot represent it will diverge from human behavior most visibly there.

\bibliographystyle{IEEEtran}
\bibliography{references}
\end{document}